\def\year{2020}\relax
\documentclass[letterpaper]{article} 
\usepackage{aaai20}  
\usepackage{times}  
\usepackage{helvet} 
\usepackage{courier}  
\usepackage[hyphens]{url}  
\usepackage{graphicx} 
\usepackage{graphicx}  
\usepackage[final]{changes}
\usepackage{dirtytalk}
\newcommand\ChangeRT[1]{\noalign{\hrule height #1}}

\title{See and Read: Detecting Depression Symptoms in Higher Education Students Using Multimodal Social Media Data}
\author{Paulo Mann,\textsuperscript{\rm 1} Aline Paes,\textsuperscript{\rm 1} Elton H. Matsushima\textsuperscript{\rm 2} \\ 
\textsuperscript{\rm 1}Institute of Computing, Universidade Federal Fluminense, Brazil\\
\textsuperscript{\rm 2}Department of Psychology, Universidade Federal Fluminense, Brazil\\
\{paulomann@id, alinepaes@ic\}.uff.br, eh.matsushima@gmail.com 
}
\begin{document}
\maketitle

\begin{abstract}
Mental disorders such as depression and anxiety have been increasing at alarming rates in the worldwide population. Notably, the major depressive disorder has become a common problem among higher education students, aggravated, and maybe even occasioned, by the academic pressures they must face. While the reasons for this alarming situation remain unclear (although widely investigated), the student already facing this problem must receive treatment. To that, it is first necessary to screen the symptoms. The traditional way for that is relying on clinical consultations or answering questionnaires. However, nowadays, the data shared at social media is a ubiquitous source that can be used to detect the depression symptoms even when the student is not able to afford or search for professional care. Previous works have already relied on social media data to detect depression on the general population, usually focusing on either posted images or texts or relying on metadata. In this work, we focus on detecting the severity of the depression symptoms in higher education students, by comparing deep learning to feature engineering models induced from both the pictures and their captions posted on Instagram. The experimental results show that students presenting a BDI score higher or equal than 20 can be detected with $0.92$ of recall and $0.69$ of precision in the best case, reached by a fusion model. Our findings show the potential of large-scale depression screening, which could shed light upon students at-risk.



\end{abstract}
\section{Introduction}\label{sec:intro}

Mental disorders have been alarmingly increasing in the worldwide population~\cite{world2017depression}. Individuals suffering from these problems may present a combination of abnormal thoughts, perceptions, emotions, and behavior\deleted{, and includes depression, bipolar disorder, schizophrenia, anxiety, among others}. One of the most common mental disorder is depression, globally estimated as more than 300 million cases~\cite{world2017depression}. Particularly, Brazil has the highest prevalence of Major Depressive Disorder (MDD)\footnote{In this work, we use MDD and depression interchangeably.} among South American countries, with nearly 5,8\%~\cite{world2017depression}. These cases are not only valid to the general population but have also been increasingly observed in the academic environment, where students face many challenges and stressful events endorsed by academic-related situations. Reports show that graduate students are more than six times likely to experience depression and anxiety, compared to the general population~\cite{evans2018evidence}. Furthermore, a previous study has shown a higher prevalence of MDD in undergraduate courses, with up to 28,2\% of prevalence in one of the investigated courses~\cite{de2006prevalencia}. 

However, naturally, before an individual with depression receives treatment, this disorder must be detected. Many patients do not receive an earlier depression diagnosis in consultation with general practitioners, with roughly 50\% of the cases detected~\cite{kessler2002detection,mitchell2009clinical}; even worse, individuals might not have the money, knowledge, or they may have even fear of social stigma to look out for help~\cite{andrade2014barriers,roness2005help}. Because of that, the disorder may remain undiagnosed, unrecognized, and, therefore, untreated, which may further aggravate its symptoms. Thus, although the most reliable way to screen for depression is the clinical diagnosis with psychological and psychiatry doctors, it is crucial to enhance other detection options beyond the consultation-based ones that usually follows the \textit{Diagnostic and Statistical Manual of Mental Disorders} (DSM) criteria. 

Another common way of detecting MDD is relying on questionnaires, such as the \textit{Beck's Depression Inventory} (BDI) and the \textit{Center for Epidemiological Studies Depression Scale} (CES-D)~\cite{beck1996beck,radloff1977ces}. They evaluate the severity of depression through a final score obtained from the answers given to the questionnaire. There are at least two problems related to such methods. First, these questionnaires should also be handled by professionals, and the individual with MDD may not always have access to them. Second, these criteria have been defined years ago. As the world develops and evolves, the criteria to detect MDD should also change to go along with the new technologies that impact everyday routine and behavior. 

Thus, the question that arises is if we could use regularly individual-generated data to detect depression. Notably, we want to investigate online environments such as social media, where the individual may express depression symptoms in a way different from the established DSM criteria.\deleted{Along these lines, social media such as communities, microblogs, and social networks poses as a promising environment to investigate depressive symptoms and behavior.} Several previous studies have already investigated social media features that characterize a user with depressive behavior~\cite{shen2017depression,sErnalaLBBRKC18,naslund2019exploring,jeri2019association}. Related to that, there is also a great interest in using machine learning to automatically distinguish between depressive and non-depressive users using their own generated data in the social media environment, or leveraging such sites to automatically gather features inspired by the DSM and questionnaires criteria~\cite{de2013predicting,tsugawa2015recognizing,reece2017instagram,shen2017depression} \added{(we expose some of them in Section \S 2)}. \deleted{To accomplish the detection task enriched with the features from either these paths, the vast majority of works benefit from Machine Learning techniques.} Screening depression symptoms from social media is related to the recently proposed concept of high-performance medicine~\cite{topol2019high}. In contrast with the traditional active diagnosis, when the individual seeks help after observing specific symptoms, the passive diagnosis systems inform individuals of possible disorders based on constant monitoring of their health, possibly through Machine Learning, for example.

The data shared by social media users, such as social networks, microblogs, and community networks consist mainly of texts and images. However, only a few recent works\deleted{to the best of our knowledge, the previous studies} have focused on assessing depressive symptoms from multimodal sources of data~\cite{morales2018linguistically}\deleted{the texts and images}. We believe that leveraging from both texts and images, which are the most common types of user-generated data, may help to distinguish different depressive groups, as depression symptoms may manifest through both verbal and nonverbal communication~\cite{morales-etal-2017-cross}. We briefly explain multimodal learning techniques in section \S 3. \deleted{On the other hand, these studies rely on feature engineering to collect the relevant features, which might differ in different samples, possibly because of cultural differences.}

Thus, in this work, we gather data shared by higher education students from one of the largest Brazilian Universities in a broadly used picture-oriented with captions social media, namely Instagram. Next, we adopt such data and machine learning methods to classify the severity of depression symptoms directly from the verbal and nonverbal user-provided content\deleted{through a Representation Learning methodology~\cite{lecun2015deep}}. \deleted{Instagram is a picture-oriented with captions social media, providing both verbal and nonverbal communication that can be used to indicate the possible depressive behavior of their owners.} Choosing Instagram is based on the following reason: we are mainly motivated by the need of investigating the increasing number of mental disorders cases within the academic environment; accordingly, several previous works have pointed it out as one of the most trustful and used social platform by young adults~\cite{shane2018college,huang2018motives}. 

As ground truth, we use the results of the Portuguese translation of \textit{Beck's Depression Inventory} (BDI) collected from an online, voluntarily answered, questionnaire{\footnote{We conducted the research under the approval of the ethical committee of the \textit{Universidade Federal Fluminense} (UFF), CAAE: 89859418.1.0000.5243}}. Our primary research question is whether we can induce Machine Learning models from \emph{a set of Instagram posts} that can distinguish students with moderate or severe depression symptoms from the others. Additionally, we would like to investigate if a model built from both images and texts performs better than using either only images or texts. We also want to assess whether we can achieve better results by learning the features and the classifier \emph{directly} from the shared data with representation learning models, to avoid the burden of inventing, engineering, and selecting specific metadata. Finally, to alleviate the negative black-box aspect of using representation learning methods, we also analyse the coefficients of a linear SVM over the induced features.




Our main contributions are as follows: (1) we create a methodology based on local search to generate a stratified oriented-to-the-individual dataset, with each example composed of a set of posts of a single individual (section {\textsl{Dataset Generation}}) so that our inferences do not consider only snapshots of posts but the target student instead; (2) we induce and compare the performance of several models that learn from \emph{representation learning}~\cite{lecun2015deep} techniques (section {\textsl{Deep Learning Models}}) and compare them with classifiers based on \emph{metadata features} (section {\textsl{Feature Engineering Models}}), both from textual and visual data; (3) we propose an early fusion neural network-based architecture to handle together the textual and visual features from posts\deleted{ in a similar representational space} (section {\textsl{Multimodal Classification}}). All code is available at our GitHub repository\footnote{https://github.com/paulomann/ReadAndSee}.

The obtained results point out that the deep multimodal classifier reaches precision and recall values good enough to be useful in the task of screening depression using Instagram. The feature engineering models are competitive in terms of F1 score compared to the deep learning models. However, deep learning systems naturally lead to transfer the trained weights to other related domains or tasks. Furthermore, they avoid the effort of investigating and engineering the metadata to solve the task.  Novel methods can provide further interpretability of black-box deep learning models.

\section{Detecting Depression (Symptoms) from Social Media}\label{sec:2}
Guntuku \textit{et al.} survey the two main ways of assessing depression from social media, namely (1) using answers of psychological tests as attributes to fed a supervised machine learning task; (2) extracting public social media data shared by individuals that have declared themselves as suffering from depression~\cite{guntuku2017detecting}. 
In the present work, we follow a hybrid approach: we rely on the BDI psychological test to obtain the class attribute, but the features come from the user-provided content. In this way, we have a more reliable class than the auto-declaration and, at the same time, more intrinsic and general features than the ones observed in the tests, aiming at fulfilling our goal: to investigate if there are underlying patterns from the user-provided content that may point out some depression tendency. 

Previous works have also followed such a hybrid approach to investigate the predictive characteristics of depression reflected in the content of social media. In~\cite{de2013predicting}, for example, tweets from individuals that answered the CES-D test were the content source. They created a binary supervised classification test according to a threshold of $22$ in the value of the CES-D test. However, \added{different from us that want to assess whether it is possible to avoid the effort of creating metadata by learning directly from the data}, they rely only on feature engineering to extract attributes encompassing depressive language, linguistic style, emotion words, among others. In ~\cite{tsugawa2015recognizing}, the methodology was the same as the previous work but targeting Japanese individuals recruited from an advertisement posted on Twitter. A surprising aspect observed from these both studies is that the former results have pointed out that the posting time and the numbers of followers and following are crucial attributes to distinguish between depressive individuals and the others. However, in the later, this difference was not observed, suggesting that cultural aspects, or merely the observed sample of individuals, may interfere in the detected patterns of depression. 

In \cite{shen2017depression}, the authors focus on classifying people from the general population as depressed or not based on their tweets. The positive examples were the ones satisfying the pattern \say{(I'm/
I was/ I am/ I've been) diagnosed depression}, or the ones that loosely mention \say{depress}. They build the machine learning models using features extracted from the tweets, computed from the users behavior in the social media and their profile. They create a multimodal dictionary to handle the features represented by different types (numeric, vector, \textit{etc.}). That work was later extended in \cite{shen18depression} to transfer a model learned from one social site to another one, aiming at avoiding labeling new data. All those features are enlightening and grounded in psychological theories, but here we would like to mainly investigate how deep learning classifiers performs when trained directly from the data, avoiding the efforts invested in engineering metadata.\deleted{investigate if how classifiers obtained directly from the shared data perform, to maybe replace in the future the effort invested on engineering metadata. }

A similar motivation inspired the work presented in~\cite{trotzek2018utilizing}, where convolutional neural networks are trained from linguistic metadata (gathered with \textit{Linguistic Inquiry and Word Count} (LIWC) tool and others) and from embeddings of textual content. Several different embeddings techniques were also used in~\cite{orabi2018deep} to detect depression from tweets. Different from the two later and the two previously mentioned works, we investigate the data from Instagram, which is picture-oriented, making the users express their feelings and state-of-mind using both nonverbal and verbal communication~\cite{morales-etal-2017-cross}. We build a fusion model to consider these types of data.

Regarding the nonverbal communication, in~\cite{reece2017instagram}, the authors aim at distinguishing posts of individuals with depression from the rest of the users using metadata and measures related to the published images (for example, the number of likes, number of comments, number of faces in the images, \textit{etc.}). They investigated the color patterns of the images, based on studies pointing out that individuals with depression tend to see the world more in tones of gray. We, on the other hand, also benefit from the captions of the pictures and from visual features learned directly from the pictures.

Previous works have also demonstrated that the pattern of social media usage is different among depressed and non-depressed users on both Twitter and Facebook~\cite{park2013perception,park2013activities}. In this work, however, we assess whether this pattern exists --- or not --- by leveraging Machine Learning models capable of distinguishing depressed and non-depressed behavior automatically.

Some of the previous works classify the posts in social media instead of the individuals. However, they are only short-content snapshots, due to the online communication nature, and probably do not have enough information to classify depression symptoms. For us, one example in the dataset is composed of a set of posts collected during a a certain period, in this way, we make the classification robust, and less error-prone.


\section{A Brief on Multimodal (Fusion) Learning}\label{sec:3}
Multimodal learning techniques induce a model by combining more than one modality of data, such as text, images, audio, video, \textit{etc.}, to solve applications ranging from the alignment of multiple data to classification from distinct sources~\cite{ngiam2011multimodal}. Recently, multimodal learning has increasingly gained attention due to the possibility of extracting latent features represented in a low-dimensional vector space with Deep-Representation learning~\cite{ramachandram2017deep}. Furthermore, this way of tackling data is particularly useful for the social media environment, where the users may express their feelings and thoughts using text, pictures, and even short videos~\cite{DBLP:journals/corr/abs-1708-02099}.

To leverage those different data sources to induce a single, unified model, one can either fuse the data following a feature-based approach (early-fusion) or a decision-based approach (late-fusion)~\cite{baltruvsaitis2018multimodal}. In the first case, one may extract the features for each modality separately, followed by merging the features to feed a classifier. When using Deep Learning, commonly, the feature extraction process is to collect the weights matrix of a layer in the network~\cite{ramachandram2017deep}. The other possibility, still in the feature-based approach, is to extract the features in a shared space, by jointly creating them from the multiple sources of data. In the decision-based approach, the final answer is based on the decisions taken from each modality by combining them using, for example, a voting process. 
The type of modality faced by Instagram data is particularly challenging as they are characterized by \emph{meaning multiplication}~\cite{bateman2014text}: the caption and the pictures in the same post may refer to distinct contexts, but both modalities are essential to creating a new meaning that diverges from merely making a decision separately from the unimodal meanings. To tackle that, in this work, we contribute with a model that induces a classifier from concatenated textual and visual features.

Previous works have also focused on multimodal social media data sources to detect disorders, for example, the relationship between eating disorders and the removal of posts from Instagram~\cite{chancellor2016post}. Focusing on depression, the work presented in~\cite{victor2019detecting} considers visual and verbal communication features in their dataset. The data was produced specifically to conduct the research, and not on a regular-basis data added in social media. Here, we are particularly interested in laying the foundations of a passive diagnosis from social media instead. Audiovisual features are also combined to detect depression symptoms in~\cite{scherer2014automatic}, using a dataset created from dyadic interactions between an interviewer and paid participants. In~\cite{morales2018linguistically}, several fusion approaches are built from features extracted from video, audio, and transcripts. The dataset is made through interviews conducted by an animated virtual interviewer controlled by a human in another room. In this work, we also investigate the benefits of a fusion architecture, but, different from there, from data extracted from a social media.

\section{Methods}\label{sec:methodology}
In order to induce the machine learning models, both the proposed models that learn directly from social media data and the ones based on metadata, it is first necessary to create the datasets\deleted{Our general framework comprises of the data harvesting and its transformation to fit a Machine Learning task and the induction of the ML models}. In the next subsections, we describe how we perform these major tasks, namely the data collection, the dataset generation, and the induction of ML models.

\subsection{Data Collection}
To collect the Instagram data published by the students, we first\deleted{For this study, we} created a Google forms questionnaire composed of (1) a number of\deleted{several} demographic questions\added{, such as the time spent on Facebook, Twitter, and Instagram; if they were diagnosed with depression; if they work; monthly pay income; Instagram username, \textit{etc}.,} and (2) the already mentioned psychometric test, BDI. Then, we published a call for participation\deleted{it} in various Facebook groups, and \added{also} asked the \textit{Universidade Federal Fluminense} (UFF) to publish \added{the call} through \added{the official} email lists. The volunteers were presented with a written explanation of the overall goals of the project, the information that would be gathered, and how their information would be used. To answer the questionnaire, they needed to be regularly enrolled in any course \added{of the University} and be\deleted{have} at least 18 years old\deleted{of age}; to ensure data integrity, we used the transparency portal that the University provides{\footnote{https://app.uff.br/transparencia/}} to validate the students registration number and their enrollment status. We did not have any personal contact with the students as the whole process was performed online\deleted{, and all data was collected X days prior to the day of the response to the questionnaire, where X can be 60, 212 or 365 days}. \deleted{We have not primed the students for the data collection, \textit{i.e.} we did not have any personal contact with the students as the whole process was performed online.} \deleted{Some of the demographics questions were the following: University registration number; time spent using Facebook, Twitter, and Instagram; if they were diagnosed with depression; if they work; monthly pay income; Instagram username and approval to gather data and \textit{etc}.}

We relied on BDI as a primary tool to assess the severity of the depressive symptoms in a student \added{and to annotate the examples}. BDI is a questionnaire comprised of 21 self-reported questions about the mental and psychological state of the individual, wherein each question has a score from zero to three points\deleted{associated} to determine the level of that specific symptom severity, where higher scores mean higher levels of that symptom. The final score is the sum of all the 21 questions scores\deleted{separately}. It can be interpreted as follows: 0--13, minimal; 14--19, mild; 20--28, moderate; and 29--63, severe~\cite{inventarioBeckII}. We first organize the data following these four intervals of depression intensity, yielding 37\% of the sample marked as severe; 23\% as moderate; 14\% as mild; and 26\% as minimal. However, as done in previous work~\cite{de2013predicting,shen2017depression} we separated the individuals into two classes: one comprising the students with non-intense depression symptoms (the ones scored in the minimal and mild classes) and the other one comprising the students with intense depression symptoms (the ones scored in the moderate and severe categories). In a real-world follow-up application of our method, the individuals classified in this last case would be the ones indicated to psychological treatment.  

We gathered the Instagram data that were posted \emph{prior to the day the survey had been taken} for each student, considering three different observation periods, namely 60 days, 212 days, and 365 days. For example, if a student answered the online questionnaire on October 15, considering the observation period of 60 days, we would collect all the student data ranging between August 16 to October 15. In this way, we prevent post introduced with the sole purpose of influencing the study. We choose 60 days because it was found to be the optimum period in~\cite{tsugawa2015recognizing}, whereas 365 was investigated in~\cite{de2013predicting}, and 212 is the mean between these two values.

\begin{figure*}[h]
\centering
\includegraphics[width=0.9\textwidth]{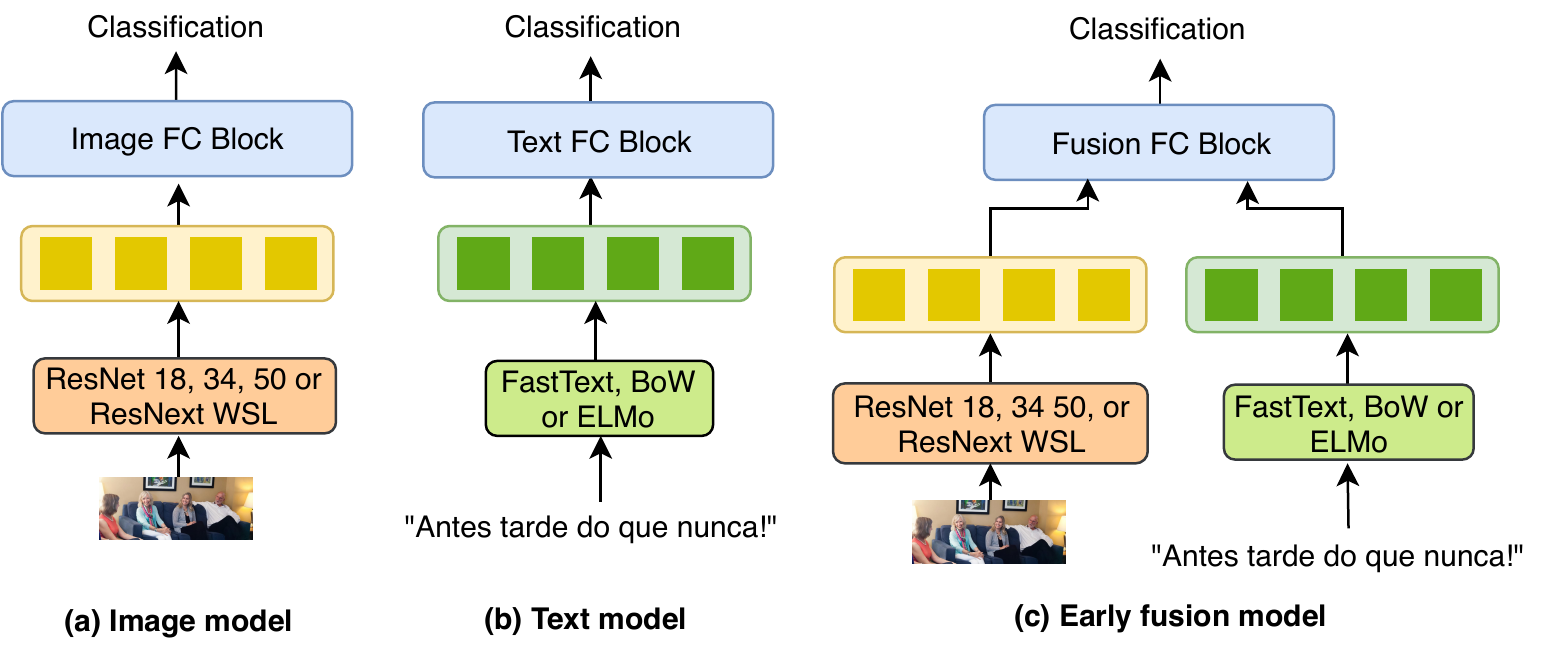}
\caption{\added{Deep learning architectures we have used to predict the intensity of depressive symptoms. Image, text, and fusion Fully Connected (FC) blocks are neural network classifiers designed especially for their particular modality.}}
\label{fig:models}
\end{figure*}

\subsection{Dataset Generation}\label{sec:data-generation}
Our target is the student classification, and not a single post, which is a snapshot of the student behavior in time. Thus, we formalized the problem as a \textit{Multiple Instance Learning} task~\cite{carbonneau2018multiple}, where the training instances are arranged in bags, and the label is provided to the entire bag. Here, the bag is the entire set of pictures or texts (or both) of each student, and the class (non-grave or grave depression symptoms) is given to the bag. In other words, the set of examples $E$ is composed of a set of bags, \textit{i.e.,} $E = \{S_1, S_2, \dots, S_m\}$, where $S_i = \{post_1, post_2, \dots, post_n\} \in E$ is the bag related to a single student $i$, and $post_k \in S_i$ is either (1) a tuple $post_k = (p_k, c_k)$ where $post_k$ is an individual post of the student, $p_k$ is a picture and $c_k$ is its caption, or (2) $post_k = p_k$, when either the post contains only a picture or when we use the examples only for image classification, or, still, (3) $post_k = c_k$, when the post is used only for text classification. Note that the size of $S_i$ may vary from student to student since we do not oblige a maximum number of collected posts. As we still need a class for each element in the bag\deleted{, Each post is annotated with the same label as the bag it belongs to, which is the student's depression symptoms classification to make our training instance (individual posts) to have a label and also to make possible to observe the impact of the possible depression symptoms over the individual state of the mind of the student. Thus}, we make each $post_k \in S_i$ to have the same label $y_i$ of $S_i$. \deleted{In this fashion, if $S_i$ has the label $y_i$ as obtained from the BDI questionnaire, we make $post_1 \in S_i = y_i, \dots, post_k \in S_i = y_i$ as well.}

\added{To acquire the training, validation, and test sets, we must require that a bag $S_i$ is not split into those different sets, as this would make the same student appearing in different phases of the learning and test process. It is also crucial to make the distribution of those sets to resemble the original distribution of the dataset. However, it is not trivial to attend all these conditions when considering both the number of bags \emph{and} the size of each bag. In this way, } to generate training, validation, and test sets, we implemented a local search method~\cite{gendreau2010handbook} to find the optimal solution in the space of candidate solutions. We start at an initial solution with three random sets $V_1$, $V_2$ and $V_3$, each one containing examples $S_i \in E$ selected at random. Next, we generate the space of candidate solutions by composing : (1) half of the solutions chosen at random; (2) for the other half, we select, at random, two bags from two distinct sets, namely, $S_{j} \in V_{w}$ and $S_{k} \in V_{p}$, and switch them making $S_{k} \in V_{w}$ and $S_{j} \in V_{p}$\deleted{that initializes with three random sets $V_1$, $V_2$ and $V_3$, each one containing examples $S_i \in E$ selected at random. To create the next population, we select at random two bags from two distinct sets, namely, $S_{j} \in V_{w}$ and $S_{k} \in V_{p}$, and switch them making $S_{k} \in V_{w}$ and $S_{j} \in V_{p}$}. The evaluation function of the local search checks if these newly generated solutions\deleted{generated sets} are better than the existing ones, according to the sum of the differences between the distributions of the new solutions\deleted{population} and the original\deleted{real} data distribution; if the new solution\deleted{population} has a better distribution than the previous best one, then the new solution becomes the selected one. The stop criteria is either the runtime (5 minutes), or when the \deleted{distribution} newly generated solution has a very similar distribution to the original distribution for the binary BDI (low intensity: $40.27\%$, high intensity: $59.73\%$)\deleted{the most similar one compared to the original distribution of posts for each BDI interval,} and to the defined dataset proportion: 60\% of the examples for the training set, 20\% for validation, and 20\% for test. After this process, we end up with ten different datasets for each observation period.

\subsection{Deep\deleted{Machine} Learning Models}\label{sec:dl-models}

Our central hypothesis is that we can build the depression classifiers directly from the data shared in the social media, avoiding the effort of building and investigating metadata. Furthermore, we argue that the meaning multiplication of multimodal data has more to add than relying only on unimodal data. To assess these assumptions, we first focus on classifiers that take the students'\deleted{We apply a number of Machine Learning models to assess our hypothesis. We first investigate Machine Learning models to classify the severity of depression symptoms in students regarding their} pictures and written posts separately. Then, we investigate how these two types of data cope together to make the final decision.

The Figure~\ref{fig:models} illustrates the three types of models examined here: (a) models created from the individual images of the students (b) models created from the individual captions; (c) a fusion model that puts together the latent features extracted from the two previous types of models. As our target is\deleted{we are interested in the classification of} the student\deleted{, and not only of the individual posts}, we combine the individual results for each post by calculating the average of all students' posts predictions to the positive class. Thus, given a student $i$ set of posts $S_i = \{post_1, post_2, ... , post_n\}$, and their respective probabilities of being in the positive class determined by the softmax function $probas_i = \{p_1, p_2, \dots, p_n\}$, we take the average of $probas_i$ to compute the student probability of being in the positive class.\deleted{taking the mean over all the probability.
Thus, given the bag related to the student $S_i = \{post_1, post_2, \dots , post_n\}$ and its corresponding ground truth label $y_i$, the classification model predicts one label for each post, given as $\hat{y}_i^{posts} = \{l_1, l_2, \dots, l_n\}$, and the voting is used to make the final bag prediction, namely $\hat{y_i}$, which is the result of the most frequent predicted labels among all the labels in $\hat{y}_i^{posts}$ list.}

\subsubsection{Image classification} To create the pictures classifier, we selected the ResNet~\cite{he2016deep} deep network as the representation learner, since it is widely used, easy to access in public frameworks, and won the ILSVRC 2015\footnote{http://image-net.org/challenges/LSVRC/2015/} competition with the ImageNet dataset. We also used the ResNeXt~\cite{xie2017aggregated} network, pretrained with Instagram images, and fine-tuned on ImageNet1k~\cite{mahajan2018exploring}, available at PyTorch Hub\footnote{https://pytorch.org/hub}. We selected this network because it was pretrained on 940 million public Instagram images, and we hypothesize that it could further help the image-based predictions. The bag associated with a single student in this case is $S_i = \{post_1, post_2, \dots , post_n\}$ and $post_k = p_k$.

We trained four\deleted{three} distinct-size architectures with the PyTorch framework~\cite{paszke2017automatic}, namely ResNet-18, ResNet-34, ResNet-50, and ResNeXt-101 32x8d, all of them starting with the pretrained weights mentioned before. To extract the latent features, we partially freeze the pretrained weights (70\%) and change the fully connected layer (FC) with the image FC block, which is a dropout layer ($p = 0.5$) followed by a linear layer.\deleted{we freeze all the pretrained weights except for the ones in the fully connected layer, which we replaced with a customized layer that includes a dropout layer ($p=0.50$) and a final linear layer with the softmax activation function We adopt the dropout layer and also the weight decay strategy to avoid overfitting since we have a relatively small training set.}\deleted{, and we also freeze all the pretrained weights of ResNets but not the closest layers to the classification layer; Since we have started from the ImageNet weights, we employ a low learning rate value of $lr = 0.0001$ to not drastically change the pretrained weights of the last layers} We induced a total of 12 image classifiers, considering the datasets created from the three observation periods (60, 212, 365), each ResNet (18, 34, 50) and ResNeXt architectures. We selected the model that reaches the best accuracy in the validation set. 

We resized the pictures to $224 \times 224$ of height and width since this is the input that both ResNet and ResNeXt implementations requires. We also standardize the pictures using the original ImageNet training mean and standard deviation. 

\subsubsection{Text classification} We use the classical Bag of Words (BOW), FastText~\cite{bojanowski2017enriching}, and ELMo~\cite{peters2018deep}\deleted{ and Doc2Vec~\cite{le2014distributed}} techniques to extract the textual feature representations. BOW is computed with SciKit Learn~\cite{scikit-learn}, FastText\deleted{ and Doc2Vec} with the Gensim implementation~\cite{rehurek_lrec}, and ELMo with the AllenNLP platform~\cite{Gardner2017AllenNLP}. In all of these cases, the examples are the captions captured from the Instagram posts, such that $S_i = \{post_1, post_2, \dots , post_n\}$ and $post_k = c_k$. If the $post_k$ has no caption, we use an empty string ($c_k = $ \say{}). After extracting the textual features with each technique, we use a text FC block, which is a linear layer, followed by a batch normalization layer, a ReLU non-linearity, and a final linear classification layer. This architecture was chosen after achieving better convergence speed in the development set.

The Bag of Words (BOW) model works by computing a value for each distinct word in a corpus. Here, our final matrix of examples when using BOW has the dimension $\sum_{i=1}^{|E|} |S_i| \times |V|$, where $|V|$ is the vocabulary size. We used the \textit{Term frequency-inverse document frequency} (tf-idf) metric to compute the value associated with each word within the example to balance the importance between frequent and uncommon terms.

Different from the BOW approach, word embeddings has a crucial role in deep learning techniques. To that end, Word2Vec~\cite{mikolov2013distributed} was one of the pioneer techniques to achieve improvements in several NLP tasks by allowing words to capture multiples degrees of meaning through their low-dimensional latent representation. However, this technique has a few limitations that the other recent ones, used in this work, does not have. First, it can not represent polysemy because of the same vector representation for the word regardless of context. Second, all embeddings are trained to an entire corpus, which means that words not seen during training are not represented at test time. Third, it does not consider hierarchical representation for words, impairing the representation of syntax and semantics aspects.

The techniques used in this work, namely, FastText and ELMo, partially or integrally solve those limitations. FastText is similar to Word2Vec, but it is robust to noisy data, as it considers subword information, which means that it can derive representations of words from morphemes, and retrieve good representations even for a small dataset~\cite{bojanowski2017enriching}. Furthermore, it is even capable of representing some of out-of-vocabulary (OOV) words --- if their morphemes are available in training time.

ELMo, on the other hand, is a Language Model (LM), different from Word2Vec and FastText. ELMo can model polysemy, subword information with character convolutions in the first layer, and hierarchical representation with two bidirectional LSTM layers on the top. The first LSTM layer usually models aspects of syntax, while the second LSTM layer retrieves aspects of contextual meaning~\cite{peters2018deep}. The final ELMo representation layer ($ELMo^{task}_{k}$) is generated by a linear combination of all these layers, which are softmax-normalized. By relying on ELMo, we allow for the implicit capture of syntax and context-dependence aspects, leaving to the model to decide which one is the most important to the task of screening depression.

Given that we were only able to collect a small dataset, we used pretrained Portuguese weights for both models: FastText weights as provided by Facebook\footnote{https://fasttext.cc/docs/en/crawl-vectors.html}, and ELMo weights by AllenNLP\footnote{https://allennlp.org/elmo}, both pretrained on a dump of the Portuguese Wikipedia. Moreover, since ELMo and FastText retrieve word embeddings, we take the arithmetic mean of the word embeddings for the caption representation.

We normalize all captions by removing punctuations, emojis and hashtags. We also changed irregular entities to a specific label: we convert numbers to \say{0}, any URL to \say{url}, @username to \say{username} (since it is not a Portuguese word), and email to \say{email} labels. The \added{general} architecture of the text classification model can be seen in Figure~\ref{fig:models}b.

\subsubsection{Multimodal\deleted{Image and text} classification} To classify the severity of depression symptoms using both the pictures and captions from users' posts, we define $post_k = (p_k, c_k)$, and, as in the text classification, we use an empty string if the picture $p_k$ has no caption. To obtain the multimodal features, we first retrieve the textual and the visual features according to the previous explained models. Inspired by the concept of meaning multiplication, where both picture and caption can create a new complex meaning, we concatenate the features from both modalities, and then we perform the final classification with the fusion FC Block, which is a dropout layer ($p = 0.5$) followed by a final linear layer. We only optimize the fusion FC block.

\begin{table*}[ht!]
\centering
\caption{The ten most commonly used hashtags by different groups of BDI, from the most (top) to less frequent (bottom). *Nikiti is a nickname for the city of Niterói.}
\begin{tabular}{l|l|l|l}
\ChangeRT{1.6pt}
minimal             & mild                  & moderate         & severe        \\
\ChangeRT{1.6pt}
\#art               & \#destinyrj           & \#rj            & \#love        \\
\#photooftheday     & \#womansolar         & \#erasmusstudent & \#rj          \\
\#photography       & \#inktober            & \#uffabroads  & \#tbt         \\
\#tbt               & \#inktober2018        & \#eurotrip       & \#smile       \\
\#artsy             & \#tbt                 & \#instadesign    & \#summer      \\
\#drawing           & \#photooftheday       & \#erasmus        & \#nature      \\
\#vsco              & \#pictureoftheday     & \#europe         & \#friends     \\
\#painting          & \#homesweetocean          & \#lisbon         & \#nikiti*      \\
\#artistoninstagram & \#guidetoniterói       & \#city           & \#photography \\
\#blackandwhite     & \#proudtobeofniterói & \#life           & \#mumbling  \\

\ChangeRT{1.6pt}
\end{tabular}
\label{table:hashtags}
\end{table*}

\subsection{Feature Engineering Models}\label{sec:femodels}

To compare our findings with baseline classifiers based on metadata, we also performed a feature engineering task from both modalities. We trained the machine learning models with the same three observation periods, and text pre-processing as used in the deep learning methods.

For textual features, we use the \textit{Linguistic Inquiry Word Count} (LIWC)~\cite{pennebaker2001linguistic} Portuguese translation~\cite{balage2013evaluation}, that was extensively investigated as useful to the task of detecting depression~\cite{morales2018linguistically,de2013predicting,resnik2015beyond}. LIWC is a text analysis program that counts words in psychologically meaningful categories~\cite{tausczik2010psychological}. Its words categories range from, for example, linguistic style usage, as the number of used pronouns, verbs, and adverbs; and other emotional categories such as positive and negative affect words. To obtain the user-level features, we aggregate the features over all posts by taking the arithmetic mean, standard deviation, and total sum, resulting in 64 features.

As the color of images is one of the most notable features to the human eye, we extract HSV --- hue, saturation and value (or brightness) --- features by taking the average of the pixels in the image. Furthermore, other studies found the HSV values to be correlated with the severity of depression~\cite{reece2017instagram}. We also capture the number of faces for each image using a deep-learning-based face detection model\footnote{https://github.com/ageitgey/face\_recognition}. The user-level visual features are also aggregated in the same way as the textual features, resulting in 12 features.

To evaluate the hypothesis of meaning multiplication, we also investigate the multimodality vs. unimodality by simply concatenating the above features. Different from the deep learning models, here we already obtain user-level features by aggregating each post features values. For the classification, we used the same neural network architecture as in the text FC block.

\begin{table}[]
\centering
\caption{Instagram data distribution (percentage of posts) for each observation period, and for each level of depression as obtained by the BDI.}
\fontsize{9.0pt}{10.0pt}\selectfont
\begin{tabular}{l|l|l|l|l}
\ChangeRT{1.6pt}
\textbf{Period\textbackslash BDI} & \textbf{Minimal} & \textbf{Mild} & \textbf{Moderate} & \textbf{Severe} \\ \ChangeRT{1.6pt}
60 days                                                   & 26.62\%           & 13.66\%           & 18.02\%           & 41.70\%           \\ \hline
212 days                                                   & 25.43\%           & 14.96\%           & 16.44\% & 43.17\%           \\ \hline
365 days                                                   & 26.05\%          & 14.80\%           & 15.35\%           & 43.80\%           \\ \ChangeRT{1.6pt}
\end{tabular}
\label{table:imagesDistribution}
\end{table}

\begin{table}[]
\centering
\caption{Mean and standard deviation of posts for each observation period considered in the study.}
\begin{tabular}{l|l|l}
\ChangeRT{1pt} & Mean  & Std   \\ \ChangeRT{1pt}
Posts per person (60 days)  & 16.73  & 24.67  \\
Posts per person (212 days)& 26.27 & 34.85 \\
Posts per person (365 days) & 37.04 & 46.61 \\
\ChangeRT{1.3pt}
\end{tabular}
\label{table:picturesPerPerson}
\end{table}

\begin{figure*}[h]
\centering
\includegraphics[width=0.9\textwidth]{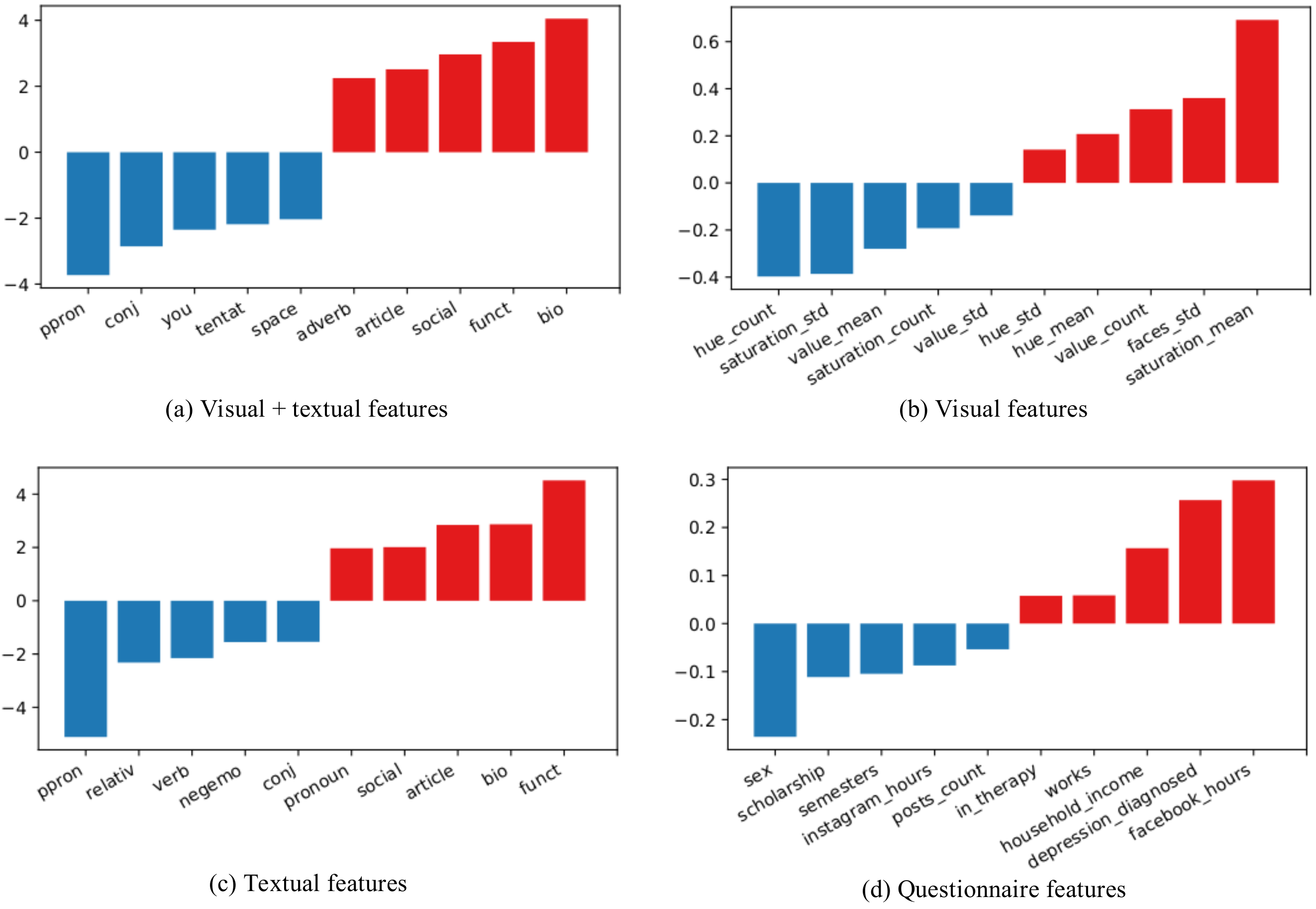}
\caption{Linear SVM's coefficient weights for predicting the positive (red) and negative (blue) classes.}
\label{figure:linearsvm} 
\end{figure*}

\section{Results}\label{sec:results}
In this section, we present the experimental results obtained from the deep learning and feature engineering models, as explained before. We start by presenting the statistics related to the student sample we gathered, followed by the results considering the demographic data, and the engineered features. To that, we inspect the coefficients weights of a linear SVM model. Next, we evaluate the classifiers on the task of screening depressed individuals using text only, image only, and both types of media. The experiments were conducted on an NVIDIA DGX-1.

\subsection{Data Statistics}
We received a total of 416 answers between October 12 and December 2, 2018, and 2--9 April, 2019. We removed six answers that were not from currently enrolled students, and 221 students agreed to provide access to their Instagram data. Thus, we have collected these 221 students data using an Instagram scraper API\footnote{https://github.com/rarcega/instagram-scraper} for Python.

Our final sample contains 136 females and 85 males with a median age of 23. For the education levels, we have 12 enrolled in Doctor's degree, 11 in Master's degree, and 198 in Bachelor's degree. For the BDI scores, we obtained a total of 82 students in the severe class, 50 in the moderate, 32 in the mild, and 57 in the minimal. We believe that the greater number of students in the severe group is because students with perceived depression might tend to participate more than their counterparts.

The Table \ref{table:imagesDistribution} shows the distribution of posts according to each category in the BDI. As we can see, students in the severe category have almost half of the data (Instagram posts) collected for each observation period considered. We can also observe in the Table \ref{table:picturesPerPerson} the mean and standard deviations of posted pictures for each observation period.

We also investigated the most frequent hashtags that the sample of students use. As we can see in Table~\ref{table:hashtags}, the mild group uses hashtags that refer to the university's city (Niterói), and state (Rio de Janeiro | RJ), where the University (UFF) is placed. On the other side, the students in the moderate group --- who could be considered as depressed --- use more hashtags related to traveling abroad\deleted{, probably a will of escaping from their current living place}. For example, Erasmus stands for European Community Action Scheme for the Mobility of University Student{\footnote{https://www.erasmusprogramme.com/}} and is a European Union student exchange program. In this group, we also have mentions to \say{\#eurotrip}, \say{\#lisbon}, and \say{\#europe}. We found intriguing the presence of so many references for traveling abroad or going to a foreign University. They might indicate a hope of a better life in another place, different from the one they are immersed.\deleted{The university name also appears in this group, which is not surprising as the sample is composed mostly of undergraduate students.} The severe group, however, was surprising as it frequently contains hashtags related to nature, summer, smile, and love. We hypothesize that the severe group might use such hashtags as a defense mechanism to alleviate depression symptoms, using a positive thinking perspective. The minimal BDI group, unlike the moderate and severe groups, focus on photography and art in general, more similar to the mild group. However, all those hypotheses require further investigation preferably conducted by a domain expert.

\subsection{Predictive Results}

We now focus on the predictive results obtained from the ML models, considering the students with the most severe symptoms as the positive class. When screening depression, it is particularly important to evaluate whether a person with high severity symptoms is incorrectly classified as possessing low severity symptoms (False Negative). Although the opposite is also important (False Positive), when screening individuals with depression, the false negative spectrum is alarming because a person with high severity symptoms, who should be detected for further treatment, is kept unknown. To that end, we choose precision, recall, and F1 metrics for model evaluation; in that way, we can have a precise measurement of how well our model is screening individuals at risk.

\begin{figure*}[ht]
\centering
\includegraphics[width=0.7\textwidth]{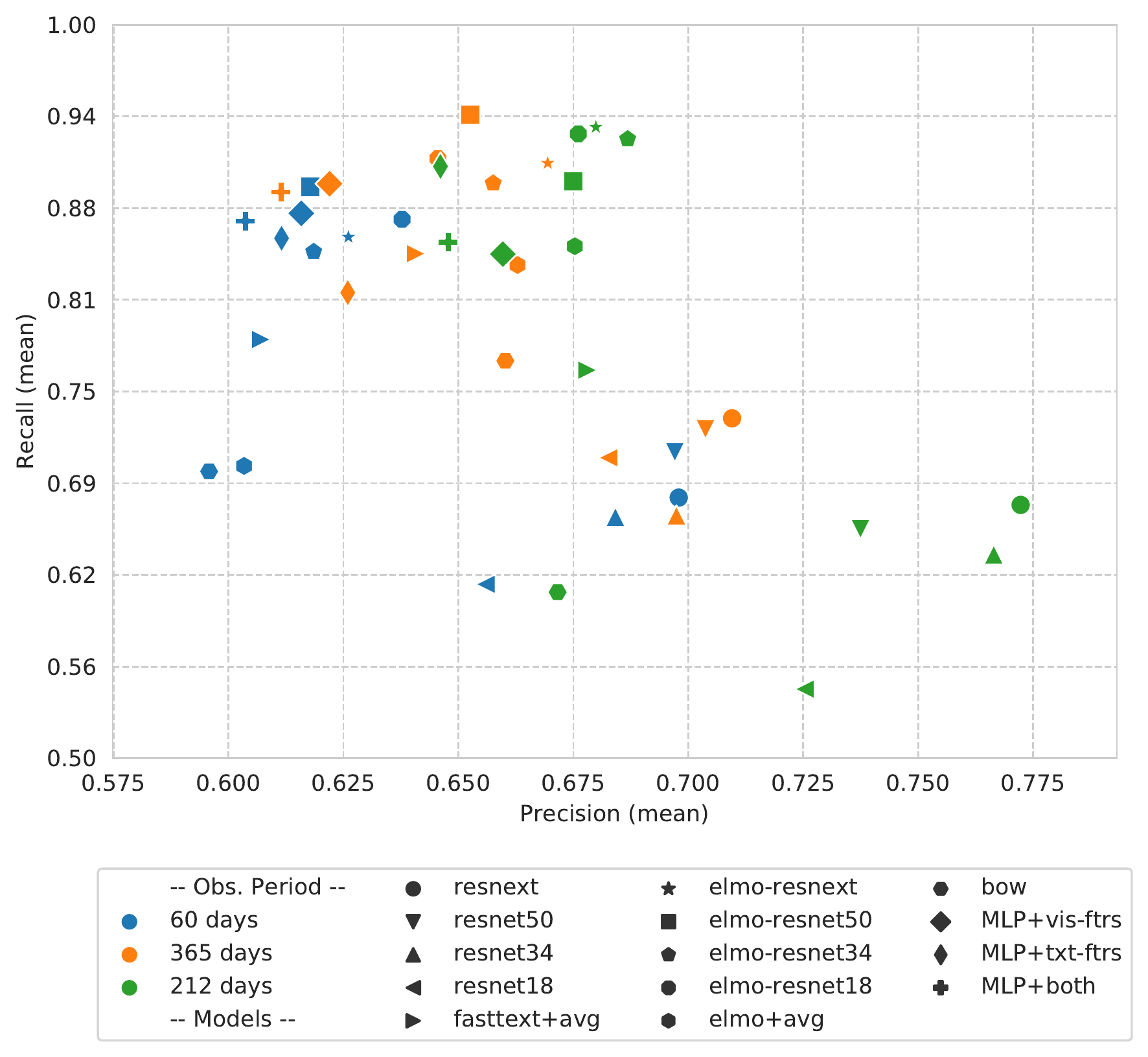}
\caption{Predictive results of the positive class using various models with different observation periods. All results are for students predictions, not posts, over 10 different datasets.}
\label{fig:allModelsResults}
\end{figure*}

We perform a 10-fold cross-validation over all experiments, and report the average metrics across all the folds. We train all models with the SGD optimizer. Table~\ref{table:hyperparameters} brings the other hyperparameters used for training. Next, we first show the most important features with the linear SVM coefficients; then, we show models' predictions results.

\subsubsection{Analysis about the sample and elicited features}

To gain insights about the classification, we employ an analysis based on linear SVM coefficients using the elicited features. We plot the top five most contributing features for the task of screening depression in Figure~\ref{figure:linearsvm}. The absolute size difference to each other can be used to determine the feature importance.

\begin{table}[h]
\centering
\caption{Hyperparameters used in the learning process. *The number of MLP hidden units is always half of the input features when not used for classification.}
\fontsize{9.0pt}{10.0pt}\selectfont
\begin{tabular}{lc|ll}
\ChangeRT{1.6pt}
\multicolumn{1}{c}{\textbf{Name}} & \textbf{Value} & \multicolumn{1}{c}{\textbf{Name}} & \multicolumn{1}{c}{\textbf{Value}} \\
\ChangeRT{1.6pt}
Epochs                            & 30             & \# MLP $h$ units               & $\frac{size(input)}{2}$*       \\
Learning rate                     & 0.001          & Batch size                        & 32                                 \\
LR decay gamma                    & 0.85           & Nest. moment.                 & 0.9 \\

LR decay epochs & 7 & Optimizer & SGD\\

\ChangeRT{1.6pt}
\end{tabular}
\label{table:hyperparameters}
\end{table}

As we can see from Figure~\ref{figure:linearsvm}c, among the most important features for classifying depression, the number of pronouns, social words --- about family, and friends ---, and bio (biological processes: eat, blood, pain) were amongst the top five correlated features for the depressed class. On the other hand, the least depressed group was correlated with the usage of personal pronouns (ppron). Although different from previous studies that found correlated signals between personal pronouns usage and depression~\cite{rude2004language,morales2018linguistically,de2013predicting}, our sample may use language differently. Particularly because in Portuguese it is not mandatory to use personal pronouns (for example, it is correct, although colloquial, to say {\textsl{\say{going to somewhere}}} instead of {\textsl{\say{I'm going to somewhere}}}). This simple example reinforces that the origin of our data may differ significantly from the previous studies, and the use of language can change across different domains.

For the visual features (Figure~\ref{figure:linearsvm}b), we found that the standard deviation of the number of faces (\say{faces\_std}), and saturation were the most correlated features with the depressed class. We hypothesize that the standard deviation of the number of faces can be correlated with depression in the sense that more depressed people post pictures, sporadically, with a higher number of friends, but not frequently. For example, they might regularly post \say{selfies,} or photographs of landscapes, and only a few pictures with a group of friends.

We also found that sex, and possessing a scholarship are correlated with the less depressed class (Figure~\ref{figure:linearsvm}d). On the other side, the time spent using facebook, total monthly income (\say{household\_income}), and whether the person was diagnosed with depression are all strongly correlated with the depressed class.

Surprisingly, when putting together both visual and textual features (Figure~\ref{figure:linearsvm}a), the results are almost the same as when using only textual features. This finding also supports previous research~\cite{morales2018linguistically,shen2017depression} that merely concatenating the values of the features do not work very well when detecting depression.

\subsubsection{Models Predictions}

We exhibit the results with a scatter plot in the Figure~\ref{fig:allModelsResults}. As one can see, results using an observation period of 60 days generally yields lower precision, along with higher recall scores. In this period, the model needs to give a \say{diagnosis} using data from 60 days only. For comparison, in a clinical setting, psychologists are encouraged to make a longitudinal evaluation, and a few sessions are not sufficient to make a final judgment, even in the presence of more evidence to support their hypotheses --- like facial expressions, hand gestures, and general body language. Thus, when we train the model with an observation period of 60 days, higher recall scores suggests that the model has sufficient information to not classify a positive as a negative example comparable with higher observation periods. We expect this behavior since the BDI questionnaire asks how respondents have been feeling during the past two weeks onset of answering the questionnaire. By this means, the model supports finding individuals at higher risk as according to the BDI, even when using less data.

On the other hand, lower values of precision suggest that the model is more susceptible to classify negatives examples as positives, which might happen due to the small number of examples for training. When we feed more data to the model, it becomes clear that there is a tendency for achieving better precision scores --- keeping, or even increasing the recall. However, there is one exception: visual-oriented deep learning models tend to have higher precision scores, even when facing only 60 days of data. This might happen because Instagram is a picture-oriented social media, and it can be easier to classify examples as true negatives using image embeddings.

For the textual representations, Bag of Words performed poorly in all settings. We hypothesize that the frequency of words, although important, is not the single most relevant feature to the task of screening depression. Previous studies have pointed out the relationship between depression and syntax, or semantics~\cite{morales2018linguistically,de2013predicting}, where ELMo has been demonstrated to leverage these features~\cite{peters2018deep}. By regarding these aspects, ELMo achieves better results compared to all the textual techniques used in this study, with nearly $0.0256$ of F1 improvement over the best FastText result. However, it is important to note that FastText is considerably more straightforward, and it is fast to train with few resources compared to ELMo.

\begin{figure*}[ht]
\centering
\includegraphics[width=0.8\textwidth]{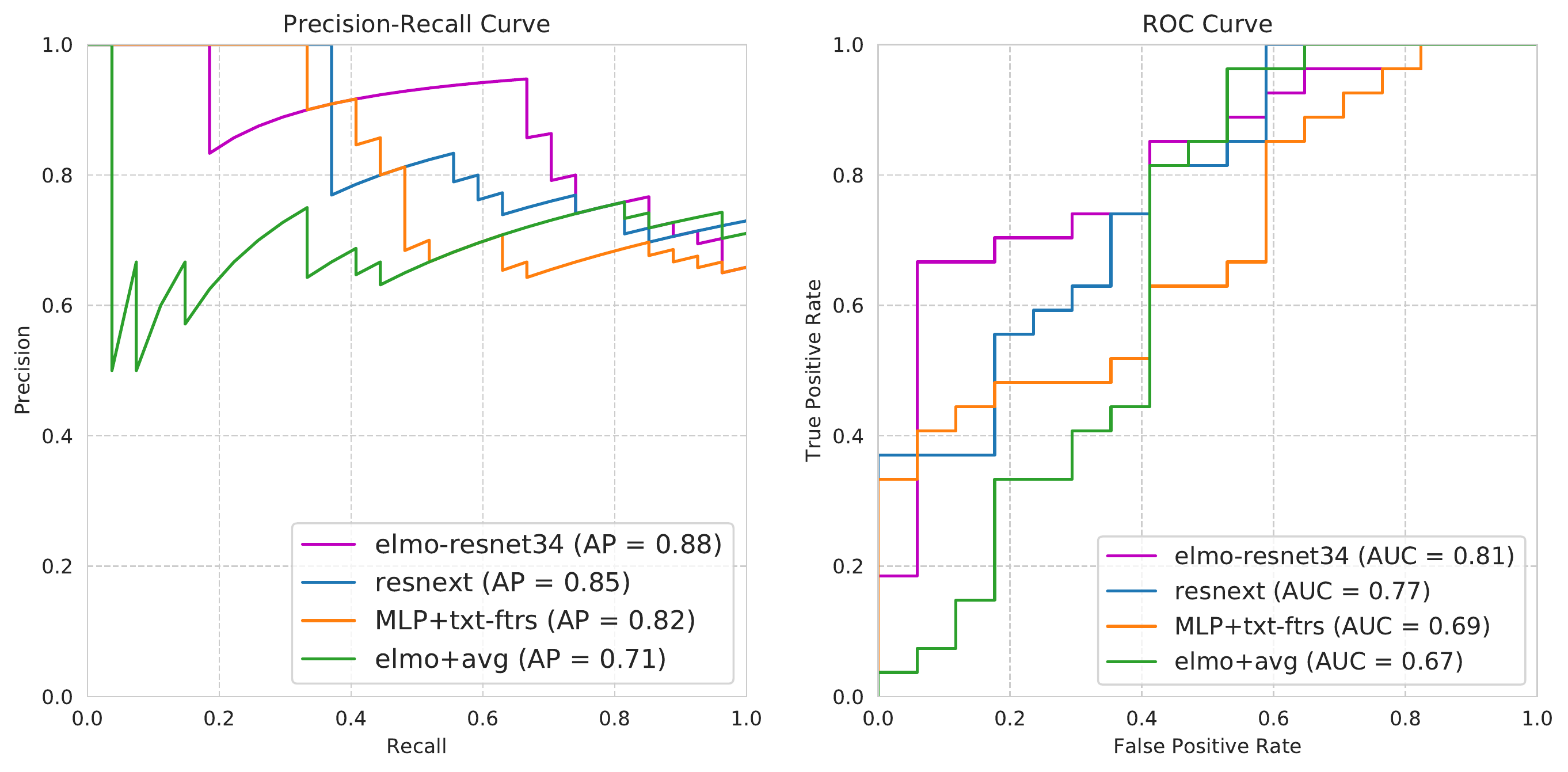}
\caption{Precision-Recall and ROC curves for the best image classifier (ResNeXt), text classifier (ELMo) and fusion classifier (ELMo + ResNet-34) for the observation period of 212 days.}
\label{fig:bestResults}
\end{figure*}

\begin{table}[h]
\centering
\caption{Best F1 results for each modality. All results are for the observation period of 212 days.}
\fontsize{9.0pt}{10.0pt}\selectfont
\begin{tabular}{lcccll}
\ChangeRT{1.6pt}
\multicolumn{1}{c}{\textbf{Model}} & \textbf{Precision} & \textbf{Recall} & \textbf{F1} & \textbf{Architecture} \\
\ChangeRT{1.6pt}
Multimodal    & 0.69               & 0.92            & 0.79        & ELMo+RN34                          \\
Text                               & 0.68               & 0.85            & 0.75        & ELMo                               \\
Image                              & 0.77               & 0.67            & 0.72        & ResNeXt                           \\
Feature Eng.                       & 0.65               & 0.90            & 0.75        & Txt features            \\
\ChangeRT{1.6pt}
\end{tabular}
\label{table:best-results}
\end{table}

Textual models usually performed better than visual models in terms of F1 score. For example, the best textual and visual models are, respectively, ELMo with $0.75$, and ResNeXt with $0.72$ of F1 scores for 212 days, as can be seen in Table~\ref{table:best-results}. The best visual result from ResNeXt is not surprising as the pretrained weights were trained with $940$ million Instagram images. However, visual models, as previously discussed, usually provided better precision scores, while textual models had higher recall scores.

For the feature engineering dataset, we had surprisingly good results. Isolated textual and visual features achieved, respectively, $0.75$ and $0.73$ of F1 score. This result is equivalent to their deep learning counterparts, but much more straightforward and naturally explain the classification, as we previously demonstrated with the linear SVM coefficients, which further supports the importance of syntax features for screening depression.

Considering the fused visual and textual features --- for the deep learning models ---, we achieve almost equivalent scores using ELMo concatenated with any ResNet, and ResNeXt architectures, where the best F1 score ($0.79$) was achieved with ELMo + ResNet-34. For the feature engineering dataset, however, the F1 score was not improved as expected when using fused features, resulting in a worse F1 score ($0.73$, for 212 days). This result can be related to the difference of features when concatenating both modalities, since we have 64 textual features and 12 visual features disposed into different representational spaces. It also indicates the necessity of more investigation on how to fuse modalities when using feature engineering, as previously explored in other studies~\cite{morales2018linguistically}. Finally, we also plot ROC and precision-recall curves in Figure~\ref{fig:bestResults} for a single dataset for the best results in each modality, as in Table~\ref{table:best-results}.

\subsection{Discussion} In general, using a deep multimodal classifier is beneficial for the task of screening depression. The feature engineering models (our baseline), on the other hand, yields competitive results when considering text or image separately; however, when using concatenated features, the results are worse. Previous studies have pointed the same direction for the screening depression task: simply concatenating engineered features makes the model focus on unimodal features instead of paying attention to both, that is why it is necessary to develop techniques for better multimodality representation, using, for example, \textit{informed fusion}~\cite{morales2018linguistically}. Our results also support this finding, for the feature engineering models, that concatenating visual and textual features do not improve model accuracy, as previously demonstrated by the SVM coefficients in Figure~\ref{figure:linearsvm}a, relying only on textual features. One possible reason is the difference in the representational space, where we have 64 features for text, and only 12 features for images. Some alignment might be necessary in order to appropriately take advantage of both modalities in this scenario.

Instagram is a picture-oriented social media platform. Intuitively, as one might expect, detecting depression using image features should lead to improved results compared to textual features. However, our findings suggest that --- with both deep learning and feature engineering --- textual features perform better than using image features only. We hypothesize that this is because people express their feelings more explicitly through written texts, making the problem easier for the ML models. However, this argument needs further investigation from the psychological literature. 

As we can see from the results, the feature engineering models yield competitive performance compared to the deep learning methods. However, we lose interpretability when using deep learning, which is important for trusting issues in AI-based systems. Nevertheless, deep learning naturally leads to transfer learning the trained weights, which in turn might be beneficial for detecting depression, as the acquired reliably-annotated datasets are usually quite small. Additionally, when doing feature engineering, one may find other features more relevant and change them across domains, which implicates on the need of retraining the entire model from scratch. Furthermore, social media usually implements the same paradigm: posts contain media, and media can be textual or visual. This paradigm simplifies the deployment of the same model across different social media platforms, leveraging previously acquired knowledge.

\section{Conclusions and Future Work}\label{sec:concl}

The ability to distinguish between different levels of depressive symptoms from social media is a promising path for passive diagnosis of individuals at risk\deleted{who were not detected in clinical visits}. To contribute in this direction, we leverage six different groups of ML architectures to distinguish students with intense depression symptoms from healthy students, relying on Instagram posts (containing both pictures and their captions). We create three deep learning models, and three feature engineering models, each based on the following media types: text-only, image-only, and the fusion of text+image. Among all the classifiers, we obtain the best predictive results with the deep multimodal classifier using ELMo and ResNet-34 concatenated features with $0.69$ of precision, and $0.92$ of recall scores. This finding suggests that a deep multimodal classifier is helpful in the task of screening depression using Instagram. Feature engineering-based models also achieve competitive results, with the advantage of more easily providing insights about the model prediction. Deep learning, on the other hand, allows for natural transfer learning across different domains, which may help when the sample is small.

As future directions, we first envision to investigate the possibility of transfer our learned models to evaluate students in other universities. We intend to address explainable deep multimodal learning by employing novel methods such as attention~\cite{vaswani2017attention}. We also expect to refine our model by interviewing the individuals and obtaining a ground truth defined by the experts. Finally, we plan to include data from other social media sites, such as Twitter and further investigate the multimodal learning possibilities.

To conclude, we believe that our contributions show a potential of help on passive diagnosis of depression, to shed light upon students at-risk and guide them to receive adequate treatment.

\section{Acknowledgments}

We would like to thank the Brazilian Research CNPq (Grant 421608/2018-8), FAPERJ (Grant E-26/202.914/2019) for the financial support, and \textit{Coordenação de Aperfeiçoamento de Pessoal de Nível Superior} - Brazil (CAPES) - Finance Code 001, for the scholarship granted to the first author. Additionaly, we would like to thank the reviewers for their valuable feed-back.

\fontsize{9.0pt}{10.0pt}\selectfont
\bibliography{bibliography}
\bibliographystyle{aaai}

\end{document}